\documentclass[letterpaper, 10 pt, conference]{ieeeconf}  

\IEEEoverridecommandlockouts                              

\usepackage[table]{xcolor}
\usepackage[english]{babel}
\usepackage{microtype}
\usepackage{hyperref}
\usepackage{cleveref}
\usepackage{cite}
\usepackage{booktabs}
\usepackage{textcomp}
\usepackage{csquotes}
\usepackage{catchfile}

\usepackage{pgfplots}

\usepackage{booktabs}

\usepackage{wrapfig}

\pgfplotsset{compat=1.15}
\usetikzlibrary{patterns,shapes,arrows,fit,positioning,shapes.geometric,shapes.symbols,shapes.misc,arrows.meta,decorations,math,hobby,calc}
\usepgfplotslibrary{statistics}
\pgfplotsset{select coords between index/.style 2 args={
    x filter/.code={
        \ifnum\coordindex<#1\fi
        \ifnum\coordindex>#2\fi
    }
}}

\graphicspath{{./img/}}

\title{\LARGE \bf
Feasibility and Acceptability of Remote Neuromotor Rehabilitation Interactions Using Social Robot Augmented Telepresence: A Case Study 
}

\author{Michael J. Sobrepera, \textit{Member, IEEE}$^{1}$, Vera G. Lee$^{2}$,\\ Suveer Garg$^{3}$, and Michelle J. Johnson, Ph.D., \textit{Member, IEEE}$^{4}$
\thanks{
This work was supported by the Department of Physical Medicine and Rehabilitation at the University of Pennsylvania and by the Eunice Kennedy Shriver National Institute of Child Health \& Human Development of the National Institutes of Health (NIH) under Award Number F31HD102165. The content does not necessarily represent the views of the NIH.}
\thanks{$^{1}$Michael J. Sobrepera is with the School of Engineering and Applied Sciences, Department of Mechanical Engineering and Applied Mechanics, University of Pennsylvania, Philadelphia, PA, USA
{\tt\small mjsobrep@seas.upenn.edu}}%
\thanks{$^{2}$Vera G. Lee was with the School of Engineering and Applied Sciences, Department of General Robotics, Automation, Sensing, \& Perception (GRASP),
        University of Pennsylvania, Philadelphia, PA, USA and is now with Johnson and Johnson, Raritan, New Jersey, United States
        {\tt\small veralee98@gmail.com}}%
\thanks{$^{3}$Suveer Garg was with the School of Engineering and Applied Sciences, Department of Electrical and Systems Engineering (ESE), University of Pennsylvania, Philadelphia, PA, USA and is now with Samsung Research, New York, NY, USA
{\tt\small gargsuveer@gmail.com}}%
\thanks{$^{4}$Dr.\ Michelle J. Johnson is an Associate Professor with the Department of Physical Medicine and Rehabilitation and BioEngineering. She directs the Rehab Robotics Lab (A GRASP Lab), University of Pennsylvania, Philadelphia, PA, USA {\tt\small johnmic@pennmedicine.upenn.edu}}%
}

\begin{document}

\definecolor{plot_color_1}{RGB}{27,158,119}
\definecolor{plot_color_2}{RGB}{217,95,2}
\definecolor{plot_color_3}{RGB}{117,112,179}

\maketitle
\thispagestyle{empty}
\pagestyle{empty}
\bstctlcite{IEEEexample:BSTcontrol}



\begin{abstract}
There is a growing need to deliver rehabilitation care to patients remotely. 
Long term demographic changes, geographic shortages of care providers, and now a global pandemic contribute to this need. Telepresence provides an option for delivering this care. However, telepresence using video and audio alone does not provide an interaction of the same quality as in-person. To bridge this gap, we propose the use of social robot augmented telepresence (SRAT). We have constructed a demonstration SRAT system for upper extremity rehab, in which a humanoid, with a head, body, face, and arms, is attached to a mobile telepresence system, to collaborate with the patient and clinicians as an independent social entity. The humanoid can play games with the patient and demonstrate activities.
These activities could be used both to perform assessments in support of self-directed rehab and to perform exercises. 

In this paper, we present a case series with six subjects who completed interactions with the robot, three subjects who have previously suffered a stroke and three pediatric subjects who are typically developing. Subjects performed a Simon Says activity and a target touch activity in person, using classical telepresence (CT), and using SRAT. Subjects were able to effectively work with the social robot guiding interactions and 5 of 6 rated SRAT better than CT. This study demonstrates the feasibility of SRAT and some of its benefits.
\end{abstract}

\section{INTRODUCTION}
Growing shortages of neuromotor rehabilitation clinicians~\cite{lin2015OccupationalTherapyWorkforce,zimbelman2010PhysicalTherapyWorkforce} and changing demographics present a challenge to delivering effective rehab care.
Frequent rehabilitation care is critical for convalescence from conditions such as stroke and cerebral palsy (CP), which can limit a person's ability to fully interact with the world around them. 
One option for overcoming these challenges is to provide rehab care using telepresence, i.e.\ telerehab, a subcategory of telehealth, the remote delivery of health care using telecommunications~\cite{howard2018TelehealthApplicationsOutpatients}.
Telehealth has recently seen a large increase in use as a result of the COVID-19 pandemic and related regulatory adjustments~\cite{latifi2020PerspectiveCOVID19Finally}.
However, telehealth with video or audio alone does not provide as rich of an interaction as in-person.
Adding a social robot to interact with patients during telehealth interactions may be able to improve those interactions.
Previous work has shown therapists believe adding a social robot to augment telehealth would significantly improve communication with patients, patient motivation, and patient compliance during telerehab interactions, compared to traditional telepresence~\cite{sobrepera2021PerceivedUsefulnessSocial}. 
Such a system could be deployed at primary care clinics, schools, or other locations in the community to provide access to rehabilitation care for patients who can not otherwise access it locally.
This would enable patients to receive more frequent care without the cost and difficulty of traveling to a distant center of rehab care excellence. 

\begin{wrapfigure}[23]{r}{.16\textwidth}
	\centering
	\includegraphics[width=.15\textwidth]{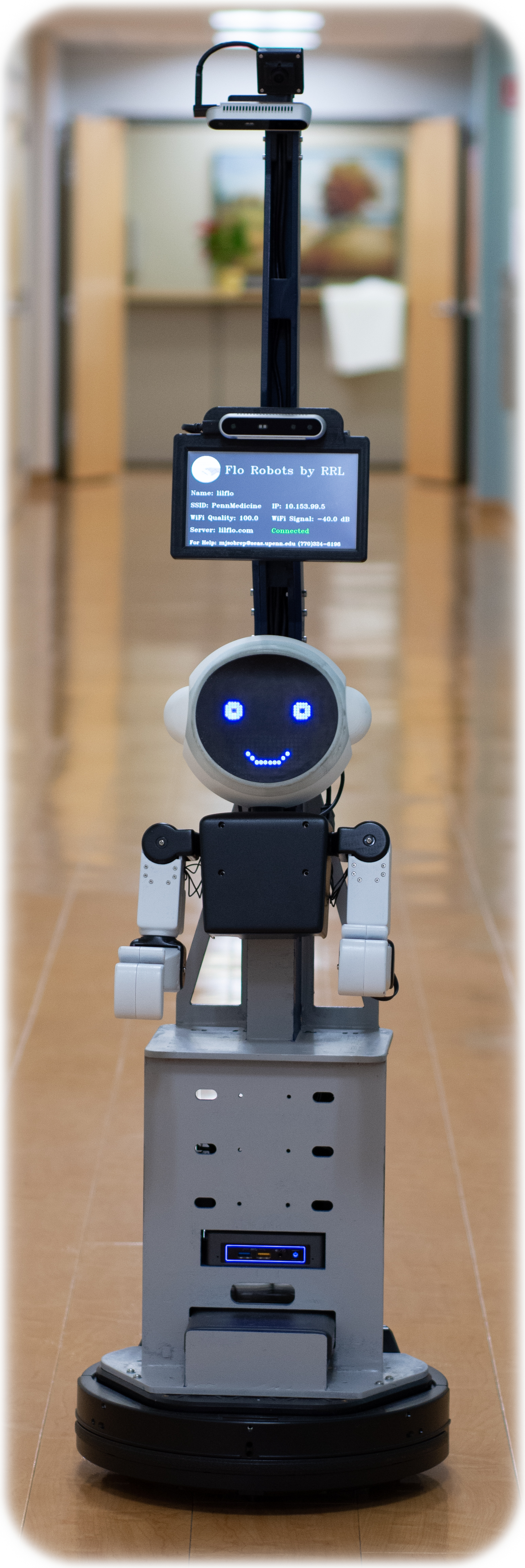}
	\caption{Flo: Social robot augmenting telepresence for remote rehabilitation.}\label{flo}
\end{wrapfigure}

Understanding how patients react to such systems will allow further development and possibly lead to better care for patients.
In this paper, we present a case series evaluating the feasibility and acceptability of social robot augmented telepresence (SRAT) for rehabilitation relevant activities, using the social robot, Flo, a small humanoid, with a torso, head, arms, and face. 
The humanoid rides on a mobile platform with a screen, cameras, and microphones (\cref{flo}).

\subsection{Related Works}
This work examines the combination of telerehabilitation and socially assistive robots.
Both have rich literature showing the promise of each for delivering rehabilitation care.

\subsubsection{Social Robots for Upper Extremity Rehab}
Social robots for rehabilitation are not new. 
They fall within the broader category of socially assistive robots~\cite{feil-seifer2005DefiningSociallyAssistive} which combine both assistive robots, which support users with disabilities, and social robots, which are designed to interact and communicate with humans.
Several systems have been developed for upper extremity rehabilitation, ex:
The NT project uses a Nao robot to interact with patients, and a Microsoft Kinect sensor to track patients, allowing the robot to autonomously demonstrate and correct poses in a pose mirroring game and in a pose sequence recall game.
In a longitudinal study of the system, with 13 subjects participating in on average 11.6 sessions, each approximately 24 minutes long, clinicians, parents, and children, found the system useful and wanted to continue to use it~\cite{pulido2019SociallyAssistiveRobotic}.
RAC CP Fun is another Nao based robotic platform designed to engage with preschool students who have CP\@ by playing games and motivating physical activity.
The robot interacts by singing songs, changing its position relative to patients, and providing feedback.
 Fridin et al.\ found that children with CP exhibited a higher level of interaction with the robot than their typically developing peers, as measured by eye contact as well as various facial, body, and vocal expressions of emotion~\cite{fridin2014RoboticsAgentCoacher,fridin2014KindergartenSocialAssistive}.
The Bandit robot is a social robot which is targeted towards older adults and is designed to monitor, instruct, evaluate, and encourage patients while they do their rehab. 
In a multi-session study with 33 subjects, the robot was found to be effective at motivating physical exercise~\cite{fasola2013SociallyAssistiveRobot}.

\subsubsection{The importance of presence in interactions}
Robots provide challenges in cost and complexity when being added to a telepresence system.
Why then are they worth pursuing?

Considerable evidence suggests that physical presence is important for motivation, compliance, understanding, and enjoyment in both rehab and non-rehab interactions.
Fridin et al.\ demonstrated with RAC CP Fun that, comparing an in-person robot and robot projected onto a screen, pediatric subjects interacted significantly more with the in-person robot~\cite{fridin2014EmbodiedRobotVirtual}.
The Bandit robot was also tested with subjects in physically present, virtually present, and simulated presence conditions. The physically present condition was preferred~\cite{fasola2013SociallyAssistiveRobot}.
Bainbridge et al.\ showed that physical presence is important for trust and motivation, especially for uncomfortable tasks~\cite{bainbridge2011BenefitsInteractionsPhysically},
and Kiesler et al.\ showed that subjects co-located with a physical robot were more engaged with it and better followed diet advice when compared to a virtual agent~\cite{kiesler2008AnthropomorphicInteractionsRobot}. 
Vasco et al.\ similarly found that stroke patients prefer using SARs for rehabilitation therapy over virtual agents shown on a screen, reporting higher engagement levels and exercise performance with the physical robot~\cite{vasco2019TrainMeStudy}
Mann et al.\ demonstrated that subjects were more likely to trust, be engaged with, and follow instructions from a robot giving instructions and asking questions, compared to the same interface on a tablet~\cite{mann2015PeopleRespondBetter}.
And C{\'e}spedes et al.\ showed that simply adding a SAR onto a pre-existing neurorehabilitation device as a third agent for motivation and engagement purposes has the potential to increase performance~\cite{cespedes2020SocialHumanRobotInteraction}.

\subsubsection{Telerehabilitation}
Telepresence systems can be as simple as using a screen, camera, and Internet connection via a cellphone, tablet, or computer, equipment which patients and providers often already possess. 
Some systems include a mobile robotic base which can be remotely controlled, such as the commercialized systems from Double Robotics and VGo Communications.
Others have robotic appendages to communicate the operator's intent~\cite{adalgeirsson2010MeBotRoboticPlatform} or screens which can actuate to face the direction the operator is looking~\cite{sirkin2011MotionAttentionKinetic,adalgeirsson2010MeBotRoboticPlatform}. 

There have been reported successes in telerehabilitation.
For example, Dodakian et al.\ presented a tabletop game system attached to a computer for rehabilitation of stroke patients.
By prompting the patient to play physical games while monitoring movements over telepresence, patient compliance and motivation was increased~\cite{dodakian2017HomeBasedTelerehabilitationProgram}. 
Abel et al.\ found that patients' range of motion can be effectively assessed over telepresence, and some patients may prefer telehealth appointments for certain rehabilitation tasks, such as range of motion assessments and wound tracking~\cite{abel2017CanTelemedicineBe}.
Preference is driven largely by convenience.
Bettger et al.\ compared a tele-physical therapy program for therapy post total knee arthroplasty (n=143) with a traditional in-person therapy program (n=144) for 12 weeks.
The telerehab program had lower costs, lower rates of rehospitalization, and was non-inferior in measures of rehab outcomes when compared to the traditional program~\cite{prvubettger2020EffectsVirtualExercise}.

However, limitations associated with this technology, including field of view of the operator (clinician), network latency, screen resolution, and projection of three-dimensional interactions into two dimensions, lessen the perception of the presence of the remote operator and reduce spatial reasoning for both users (clinician and patient)~\cite{johnson2015CanYouSee,sirkin2011MotionAttentionKinetic}.
The resulting lack of physical presence, coupled with unclear instructions for movements over telepresence, may decrease patients' compliance and motivation to perform required motor assessment tasks and, as a result, limit the clinician's ability to assess the patient's current function and progress and motivate home therapy adherence.
 This highlights a need to develop platforms that have a physical presence and can perform both assistive and social functions.
 With the current pandemic, the need and call for telerehab systems has grown~\cite{bettger2020COVID19MaintainingEssential}, a need which will continue to grow with shifting demographics and future unforeseen care challenges. 
 
 \subsection{Social Robot Augmented Telepresence}
 Recognizing the broad potential impact from telerehab and that some of the challenges with telepresence could be overcome by using a socially assistive robot, we propose to augment telerehab interactions with social robots. 
 In a previous study of 351 therapists in the United States, therapists reported they believe social robot augmented telepresence (SRAT) would significantly improve communication, motivation, and compliance during telerehab interactions, compared to classical telepresence.
 To explore this new direction in healthcare robotics, we have developed Flo, an example SRAT.
 
 The system is comprised of a Kobuki drive base, two RealSense D415 cameras, a screen, speakers, and a fisheye camera (\cref{flo}).
 A humanoid with a body, head, arms, and a face can be placed onto the system and easily removed. 
 This allows the system to be used in both classical telepresence (CT) and SRAT configurations. 
  The system is controlled through a server via a remote web interface by a plays and script methodology with wizard of oz control.
 The humanoid can synthesize arbitrary speech using AWS Polly, and the face of the humanoid can be dynamically changed~\cite{sobrepera2019DesigningEvaluatingFace}.
 More design details can be found in~\cite{sobrepera2021DesignLilFlo}. 

Here we explore the feasibility of social robot augmented telepresence for engaging in upper extremity rehabilitation activities, using the Flo platform to compare SRAT to CT. 

\section{PROCEDURE}
The study was approved by the University of Pennsylvania Institutional Review Board. 
In addition to consenting to participate in the study, all subjects also provided an optional media release, allowing publication of their images. 

\subsection{Demographics and Baselines}
After consenting, subjects were assessed using the box and block test~\cite{jongbloed-pereboom2013NormScoresBox} to measure unilateral gross manual dexterity, children's color trails test for pediatric subjects~\cite{williams1995ChildrenColorTrails} and color trails test for adult subjects~\cite{maj1993EvaluationTwoNew} to measure visual attention, graphomotor sequencing, psychomotor speed, and cognitive flexibility, as a proxy for executive cognitive function more generally, and grip strength test~\cite{mathiowetz1986GripPinchStrength} to measure hand and forearm strength as a proxy for upper limb strength (\cref{clinical_measures_exp}).

\newcommand\testimgheight{1.1in}
\begin{figure*}
    \centering
    \includegraphics[height=\testimgheight]{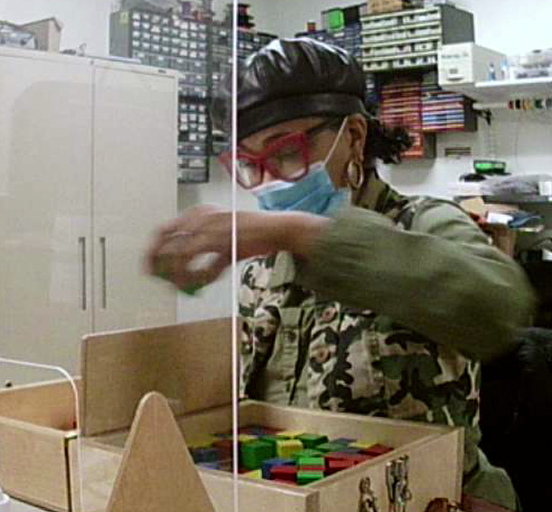}
    \includegraphics[height=\testimgheight]{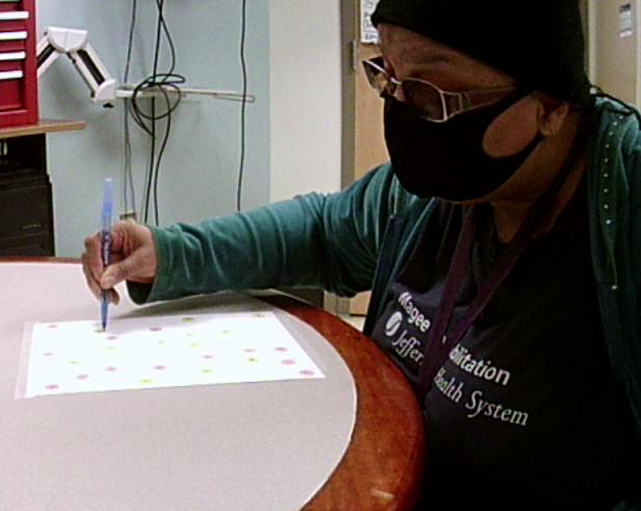}
    \includegraphics[height=\testimgheight]{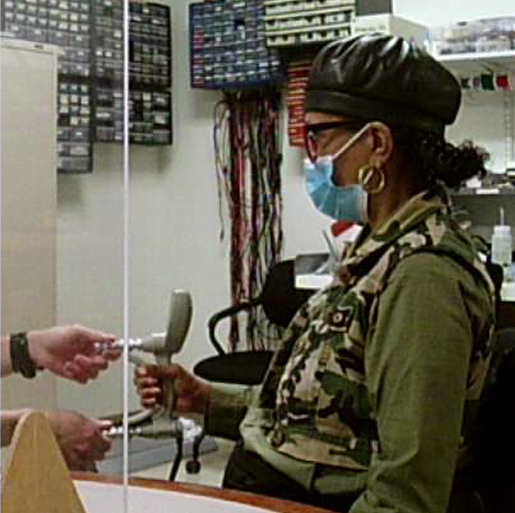}
    \caption[Subjects performing clinical assessments]{Subjects performing the clinical assessments at the beginning of the trial. From left to right: the box and block test, color trails test, and grip strength test.}\label{clinical_measures_exp}
\end{figure*}

Subjects, aided by a caretaker/parent if present, answered a  survey asking: basic demographics, history of cognitive and motor impairment, technology usage, level of education, and current therapy practice and compliance, how they are feeling today, and how they feel about robots.
Surveys were  administered by a study team member, enabling data to be collected from subjects of all ages and cognitive levels as well as allowing follow-up questions to be asked.

\subsection{Conditions and Ordering}
Subjects were randomly assigned to an ordering group.
With \textbf{FTF} signifying face-to-face (\cref{inperson_pilot}), \textbf{SRAT} (\cref{aug_pilot}), and \textbf{CT} (\cref{classic_pilot}), the groups are: FTF-SRAT-CT and FTF-CT-SRAT\@.
Once the intake surveys were complete, the subject was seated in a room.
For the FTF condition, the operator entered the room and sat on a chair, next to the telepresence system, which was not connected, but used to gather video data, in front of the subject.
For the SRAT/CT conditions, the robot entered the room under remote control by the same operator.
This ordering sequence was used to give each subject a baseline face-to-face experience to begin, as would likely happen in the real world, before moving on to the telepresence modalities.
Previous results have suggested that when initial interactions are remote, results are poor~\cite{fridin2014EmbodiedRobotVirtual}.

\newcommand\testimgheightact{1.1in}
\begin{figure*}
    \centering
    \includegraphics[height=\testimgheightact]{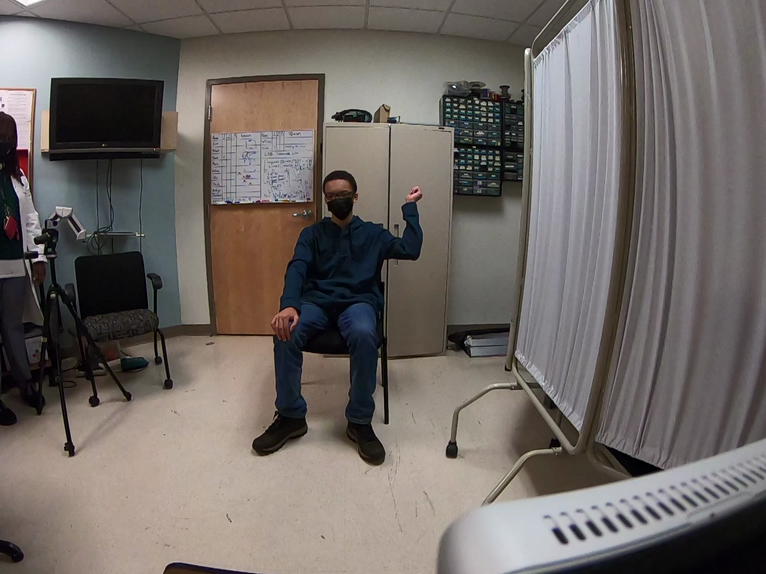}
    \includegraphics[height=\testimgheightact]{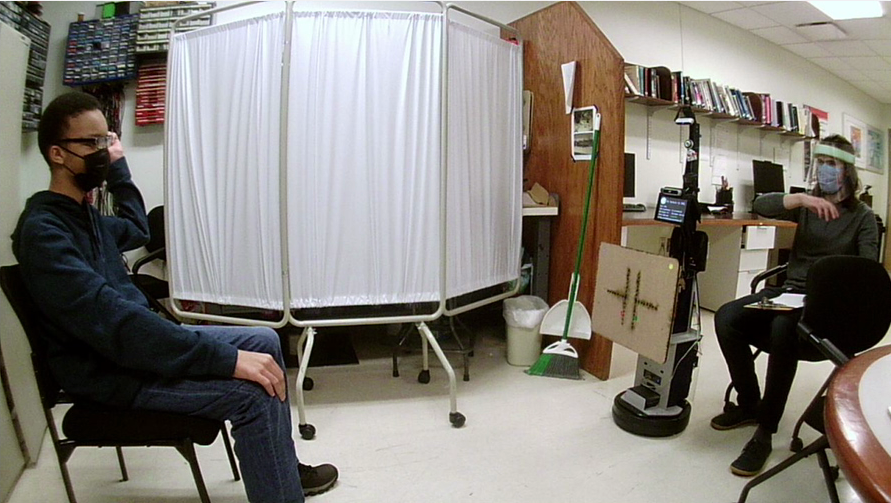}
    \includegraphics[height=\testimgheightact]{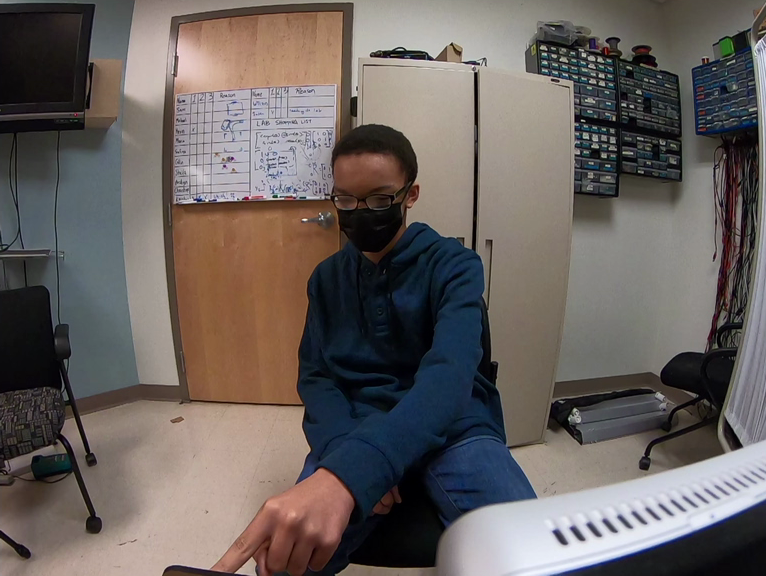}
    \includegraphics[height=\testimgheightact]{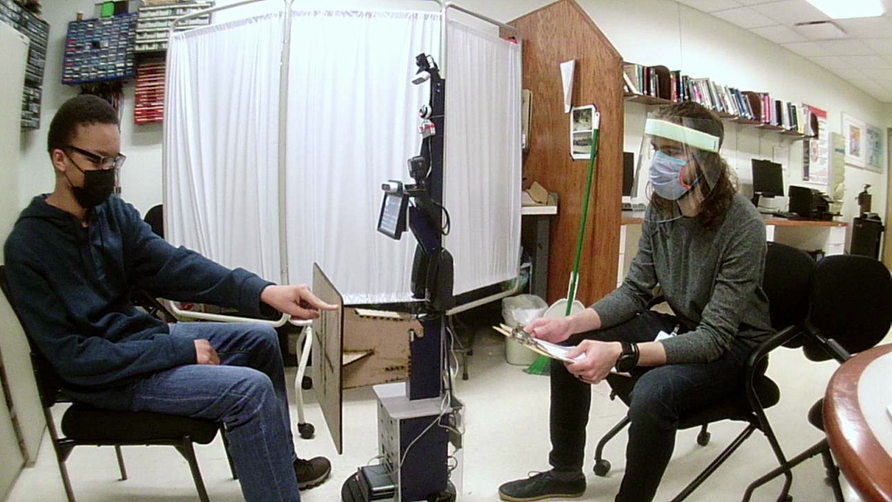}
    \caption[In person interactions during study]{Subject interacting in person with the operator (Condition FTF). On the left, playing a Simon Says game. On the right, performing the target touch activity.}\label{inperson_pilot}
\end{figure*}

\begin{figure*}
    \centering
    \includegraphics[height=\testimgheightact]{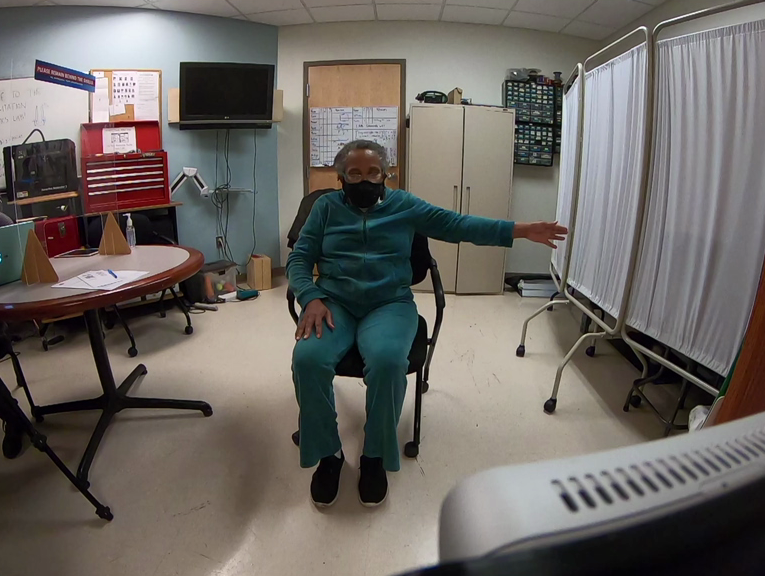}
    \includegraphics[height=\testimgheightact]{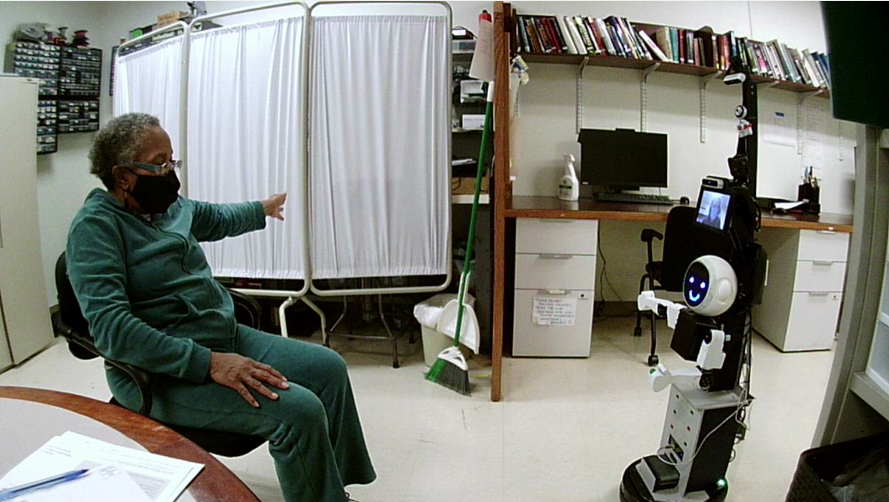}
    \includegraphics[height=\testimgheightact]{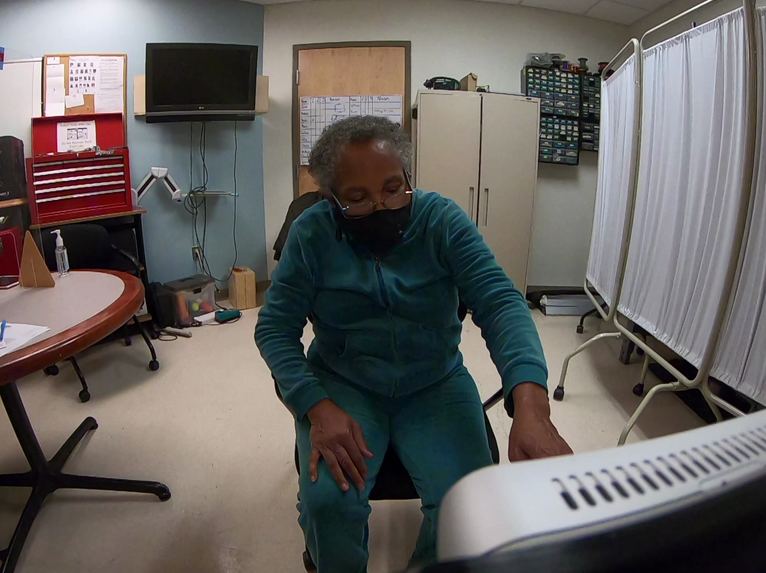}
    \includegraphics[height=\testimgheightact]{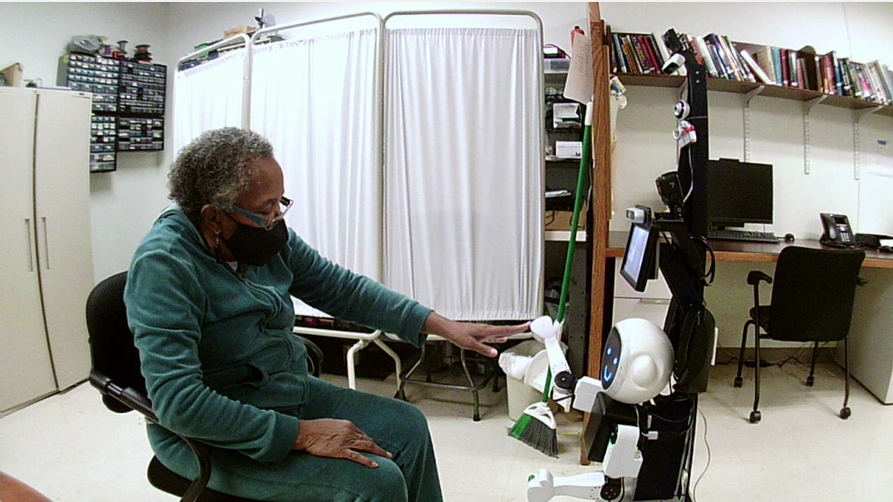}
    \caption[Interactions during study with social robot augmented telepresence]{Subject interacting with the operator via telepresence with a social robot augmenting the interaction (Condition SRAT).}\label{aug_pilot}
\end{figure*}

\begin{figure*}
    \centering
    \includegraphics[height=\testimgheightact]{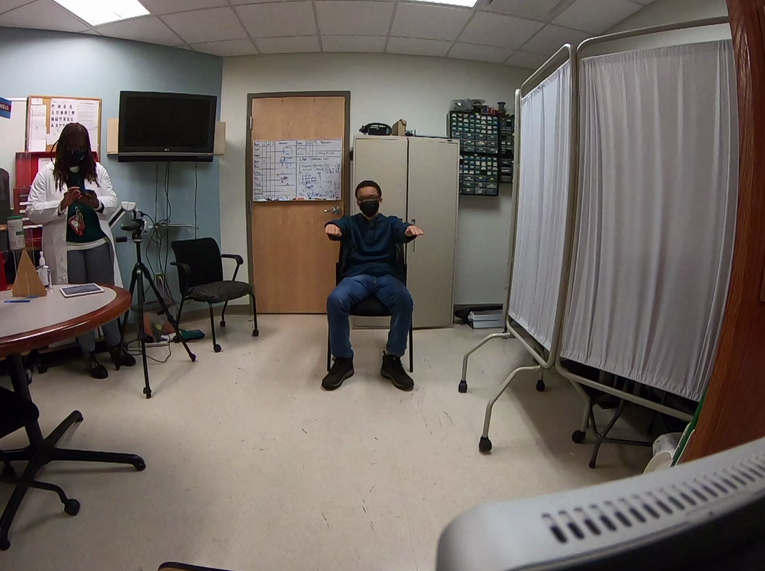}
    \includegraphics[height=\testimgheightact]{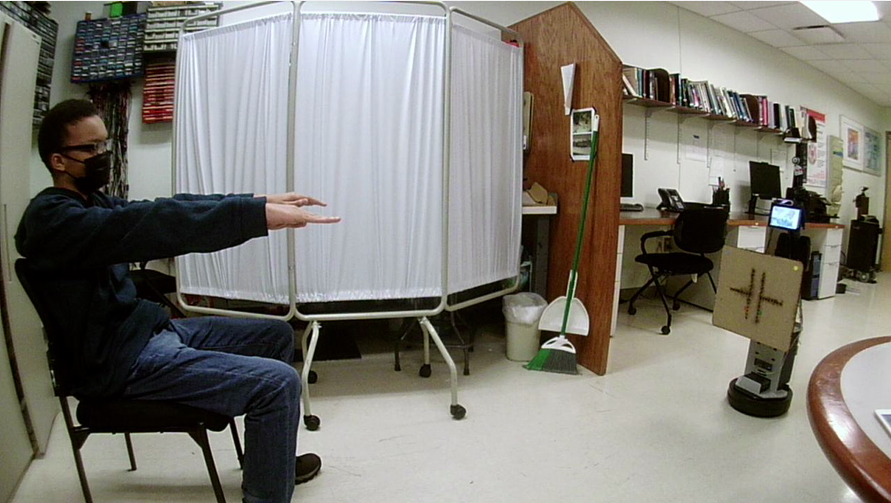}
    \includegraphics[height=\testimgheightact]{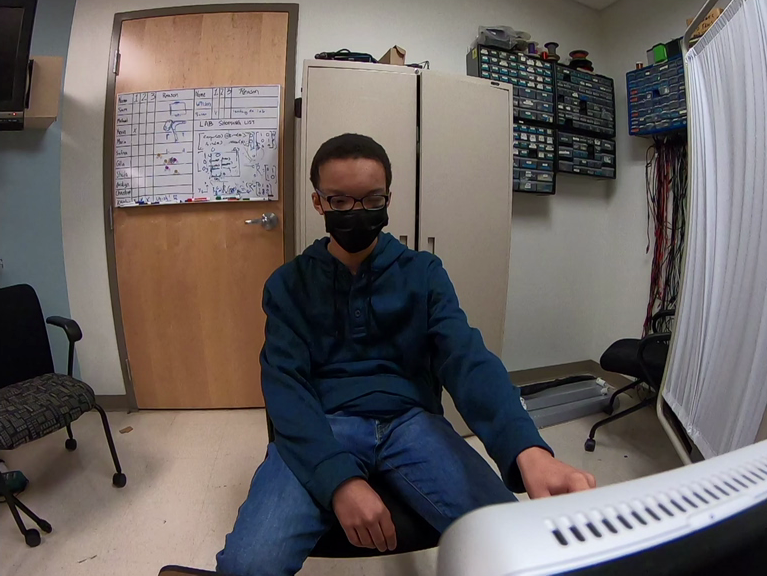}
    \includegraphics[height=\testimgheightact]{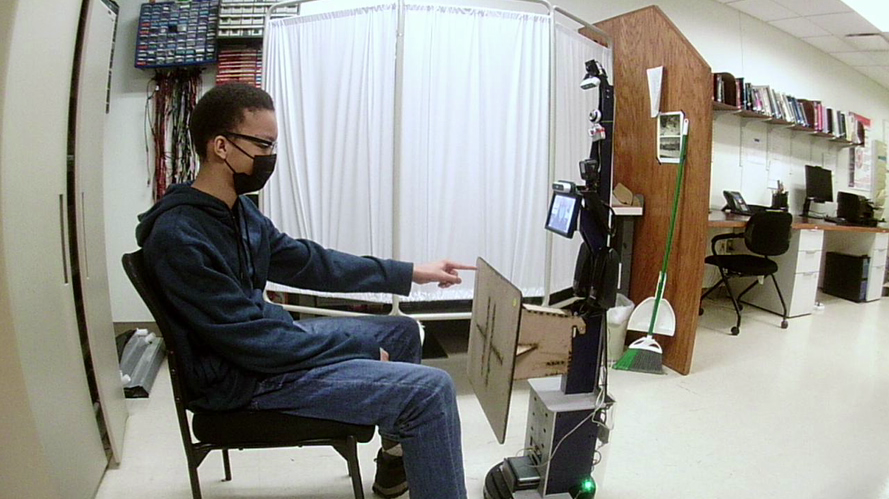}
    \caption[Interactions during study via classic telepresence]{Subject interacting with the operator via classic telepresence (Condition CT).}\label{classic_pilot}
\end{figure*}

\subsection{Interaction}
To begin each interaction, the operator introduced themselves and, in the SRAT condition, the robot.
The operator asked the subject if they wanted to play a game.
When the subject said yes: in the SRAT condition the robot said ``great, let's play Simon Says'' and explained how to play a game of Simon says, in the CT/FTF conditions, the operator said ``great, let's play Simon Says'' and explained how to play a game of Simon Says.
The game was played with the subject moving through 20 motions with an additional 6 repeated motions where there was no Simon Says command, with all instructions given by the robot in the SRAT condition.
The motions were identical in all conditions, composed of bimanual activities: clapping, reaching overhead with hands apart, reaching out forward with arms, covering both eyes and unimanual activities: reaching across the body to touch the shoulder, touching the mouth, touching the top of the head, reaching out to the side, etc.
For the first four subjects, the unimanual activities were done individually for each arm, randomly ordered. 
As the study evolved, for the final two subjects, activities were made more difficult and engaging using bilateral non-symmetrical motions constructed randomly (ex: ``wave with your right arm and touch your shoulder with your left hand'').
This game pushed the subjects to the ends of their range of motion, using motions which are relevant for activities of daily living (ADLs).

The operator then said that they want the subject to play another game.
In the CT/FTF conditions, the operator explained the game and in the SRAT condition, the robot explained it: ``In the target touch activity I will tell you to touch the dots on my [hands/board] ... Let's start in a ready position, return to this position after every touch''.
In the CT condition, the telepresence system had colored dots mounted on a board hanging on the robot.
In the FTF condition, the same board was mounted on the platform next to the operator.
In the SRAT condition, the robot had the same dots on its hands and moved its hands to the same points as those on the board.
The first four subjects were instructed to touch each of three dots with each hand ten times, for a total of sixty point-to-point motions.
The final two subjects were instructed to complete a series of 1 to 4 dot/hand sequences with two dots available, randomly selected.
This activity tests the attention and motivation of the subjects by instructing them to do repetitive motions.
In the modification made for the final two subjects, difficulty was increased, and more sustained attention was required rather than simple touching. 

\subsection{Mid-Interaction Survey and Rest}
On completion of each condition, the robot/operator exited the room.
Subjects were then given a survey asking questions from the NASA task load index (TLX)~\cite{hart1988DevelopmentNASATLXTask}, intrinsic motivation inventory (IMI)~\cite{mcauley1989PsychometricPropertiesIntrinsic}, how well they enjoyed the interaction, if they would like to do the interaction again, and how safe they felt during the interaction.
Once the surveys were completed, subjects rested until at least 15 minutes had passed since the end of the prior condition.

\subsection{Post Study Survey}
At the end of the study, subjects are asked questions about the entire experiment: which interaction modality they thought was best and several questions adapted from Telemedicine Satisfaction and Usefulness Questionnaire~\cite{bakken2006DevelopmentValidationUse} on whether they thought telemedicine would change how they manage their healthcare, communicate with their clinicians, and if telehealth visits would be convenient to them.

\section{RESULTS}
\subsection{Subjects}
Eight subjects participated in the pilot trial. 
The surveys were dramatically edited between the first and second subjects, so the first subject was excluded from analysis.
For one other subject, the power system on the robot failed, preventing that subject from completing the protocol.
As a result, that subject was also excluded. 
Therefore 6 subjects' results were analyzed.
Complete subject information can be seen in \cref{subjects}.

\begin{table*}
\centering
\caption{Subject demographics, affective state, experience/feelings regarding relevant technologies, and trial interaction preferences\label{subjects}}
\begin{tabular}{@{}llcccccc@{}}
  \toprule
&  & \textbf{AJ} & \textbf{HL} & \textbf{GS} & \textbf{VM} & \textbf{PD} & \textbf{BF} \\ 
  \midrule
  & Diagnosis & Stroke & Stroke & Stroke & None & None & None \\
  & Age & 55 & 53 & 63 & 13 & 4 & 6 \\
  & Gender & F & M & F & M & F & F\\
  \multicolumn{8}{l}{\textbf{Color Trails Test Z-score}}\\
  & CTT-1 & -3.9 & -2.2 &  0.0 & -0.8 & - & - \\ 
  & CTT-2 & -5.0 & -1.5 &  0.9 &  0.4 & - & - \\ 
  \multicolumn{8}{l}{\textbf{Box and Block Test Z-score}}\\
  & Dominant hand & -4.6 & -3.1 & -5.6 & -3.2 &  0.2 & -3.2 \\ 
  & Non-dominant hand & -2.3 & -2.3 & -5.2 & -3.5 &  0.3 & -3.1 \\ 
  \multicolumn{8}{l}{\textbf{Self-Assessment Manikin}}\\
  & Valence (1: Happy, 9: Unhappy) & 1 & 1 & 1 & 3 & 1 & 1 \\ 
  & Arousal (1: Excited, 9: Relaxed/Sleepy) & 1 & 1 & 5 & 4 & 1 & 3 \\ 
  & Dominance (1: Submissive, 9: Dominant) & 4 & 7 & 9 & 7 & 9 & 9 \\ 
  \multicolumn{8}{l}{\textbf{Please rate your level of experience with (1: No experience, 5: Very high experience):}}\\
  & Computers & 4 & 2 & 3 & 4 & 2 & 2 \\ 
  & Tablets & 4 & 2 & 3 & 4 & 3 & 3 \\ 
  & Smartphones & 4 & 2 & 2 & 4 & 3 & 1 \\ 
  & Robots & 4 & 2 & 1 & 3 & 2 & 1 \\ 
  \multicolumn{8}{l}{\textbf{Have you ever done:}}\\
  & A video call? & Yes & Yes & Yes & Yes & Yes & Yes \\ 
  & A video call for healthcare? & No & No & Yes & No & No & No \\ 
  \multicolumn{8}{l}{\textbf{How do you feel about (1: Very negative, 5: Very positive):}} \\
  & Using video calls for healthcare?  & 5 & 4 & 2 & 4 & 5 & 2 \\ 
  & Robots? & 5 & 5 & 4 & 5 & 5 & 4 \\ 
  \multicolumn{8}{l}{\textbf{Interaction Modality Preference:}}\\
  & First & CT & FTF & FTF & SRAT & SRAT & SRAT\\
  & Second & SRAT & SRAT & SRAT & FTF & CT & FTF\\
   \bottomrule
\end{tabular}
\end{table*}

The subjects with stroke varied functionally -- one had aphasia, another had spasticity and loss of function in one hand, preventing independent finger and wrist actuation. 
All but one subject (PD) performed 2 to 6 standard deviations below normal on the box and block test~\cite{jongbloed-pereboom2013NormScoresBox,mathiowetz1985AdultNormsBox}.
Subjects varied from severely impaired to typically functioning on the color trails test.
At intake, all subjects reported being happy, excited to neutral, and dominant to neutral. 
All subjects reported positive feelings towards robots. 
All subjects had previously made video calls, only one had done so for healthcare (GS). 
Subjects initially were mixed on their feelings towards using video calls for healthcare.

\subsection{Randomization}
AJ, BF completed the study in the FTF-CT-SRAT order. HL, GS, VM, PD completed it in the FTF-SRAT-CT order.

\subsection{Ability to Participate}
All subjects were able to fully participate with the robot. 
No subjects reported difficulty understanding the instructions given by the humanoid. 
AJ reported difficulty in the CT condition with seeing the operator due to the small screen size. 
The last two subjects to participate (PD and BF), who were the youngest and experienced more challenging activities, appeared to have more difficulty with the Simon Says activity and had significant difficulty with the target touch activity across conditions. 
PD was not comfortable with left and right hands and had difficulty with more than two step instructions.

\subsection{Quality of Interactions}
Using the IMI (1 low to 7 high), enjoyment for each condition averaged between 5.1 and 5.8, competence averaged between 6.5 and 6.7, effort averaged between 3.8 and 4.4, and pressure averaged between 2.1 and 2.4.

\subsection{Modality Preference}
All three pediatric subjects rated SRAT as the best modality. 
The three adult subjects rated SRAT as the second-best modality.
Complete ratings are shown in \cref{subjects}.

\subsection{Observations}
Throughout the interactions, the subjects had opportunities to provide feedback and the study staff recorded notes. 
GS reported not liking telepresence, but liked working with the social robot, and stated that they felt less pressure as the robot could adjust its pacing to their needs (compared to a human who may be in a rush). 
AJ said that they thought the social robot would be useful for home practice. 
VM reported that the SRAT interaction was the best due to it being the funniest. 
BF was more energetic filling out surveys after the SRAT interaction than the other interactions.
PD, the youngest subject, was very excited to interact with the social robot. 
After completing the experiment and surveys, they asked if they could play with the robot more.
We drove it into the room and had the robot say hello, wave, and talk with them. 
They interacted with it and ended up petting it on its head as if it were a pet or a doll. 

\section{DISCUSSION}
This case series covers users of a wide variety of ages and motor and cognitive function.
All subjects were able to fully participate in three conditions: face to face, classical telepresence, and social robot augmented telepresence. 
Enjoyment was high for all conditions.
All conditions required moderate effort, but subjects reported high competence and felt low pressure.
With this size of a case series, comparisons between measures after conditions are not warranted.
The high level of enjoyment and the ability of all subjects to interact with the social robot while it augmented the telepresence interaction demonstrates acceptability of the system by patients.

The social robot augmented telepresence system was rated higher than traditional telepresence by five of the six subjects (all but AJ).
This is an early indicator of the benefit that the social robot can add. 
The pediatric subjects all rated the social robot augmented condition as best. 
This may be partly a novelty effect, which is well documented in the literature.
But previous systems, such as the NT, have demonstrated the use of social robots for rehab longitudinally with good results~\cite{pulido2019SociallyAssistiveRobotic}.
Even if there is a large amount of novelty driven interest, by using evolving games and activities coupled with dynamic robot expressions, high engagement should be maintained over time. 
Seeing the youngest subject's (PD) high enthusiasm for the social robot was exciting to the study team. 
That subject clearly assigned a personality to the robot. 

One of the adult subjects commented on the social robot being well suited for subjects who move slower, because it can adapt to their pace, reducing the pressure during visits while encouraging improved performance. 
This highlighted a point of utility that we had not anticipated and adds more rationale for developing social robots for rehabilitation and remembering to design with elders, not just children, in mind.

\subsection{Limitations}
All subjects had positive views towards robotics at the start of the trial and had previous experience making video calls, although only one for health care purposes, which may bias the respondents towards accepting the social robot. 

\section{CONCLUSIONS}
We completed a case series with 6 subjects interacting with the Flo robotic system, an example of social robot augmented telepresence.
By multiple metrics, the system performed well, was accepted by subjects, and appeared feasible to use for upper extremity motor rehabilitation activities and assessments. 





\section*{ACKNOWLEDGMENT}
We thank the subjects who participated in this work.


\bibliographystyle{IEEEtran}
\bibliography{IEEEabrv,ieee_bib_control,refs}

\begin{thebibliography}{10}
\providecommand{\url}[1]{#1}
\csname url@samestyle\endcsname
\providecommand{\newblock}{\relax}
\providecommand{\bibinfo}[2]{#2}
\providecommand{\BIBentrySTDinterwordspacing}{\spaceskip=0pt\relax}
\providecommand{\BIBentryALTinterwordstretchfactor}{4}
\providecommand{\BIBentryALTinterwordspacing}{\spaceskip=\fontdimen2\font plus
\BIBentryALTinterwordstretchfactor\fontdimen3\font minus
  \fontdimen4\font\relax}
\providecommand{\BIBforeignlanguage}[2]{{%
\expandafter\ifx\csname l@#1\endcsname\relax
\typeout{** WARNING: IEEEtran.bst: No hyphenation pattern has been}%
\typeout{** loaded for the language `#1'. Using the pattern for}%
\typeout{** the default language instead.}%
\else
\language=\csname l@#1\endcsname
\fi
#2}}
\providecommand{\BIBdecl}{\relax}
\BIBdecl
\renewcommand{\BIBentryALTinterwordstretchfactor}{4}

\bibitem{lin2015OccupationalTherapyWorkforce}
V.~Lin, X.~Zhang, and P.~Dixon, ``Occupational {{Therapy Workforce}} in the
  {{United States}}: {{Forecasting Nationwide Shortages}},'' \emph{PM\&R},
  vol.~7, no.~9, pp. 946--954, Sep. 2015.

\bibitem{zimbelman2010PhysicalTherapyWorkforce}
J.~L. Zimbelman, S.~P. Juraschek, X.~Zhang, and V.~W.-H. Lin, ``Physical
  {{Therapy Workforce}} in the {{United States}}: {{Forecasting Nationwide
  Shortages}},'' \emph{PM\&R}, vol.~2, no.~11, pp. 1021--1029, Nov. 2010.

\bibitem{howard2018TelehealthApplicationsOutpatients}
I.~M. Howard and M.~S. Kaufman, ``Telehealth applications for outpatients with
  neuromuscular or musculoskeletal disorders: {{Telehealth Applications}} for
  {{Outpatients}},'' \emph{Muscle Nerve}, vol.~58, no.~4, pp. 475--485, Oct.
  2018.

\bibitem{latifi2020PerspectiveCOVID19Finally}
R.~Latifi and C.~R. Doarn, ``Perspective on {{COVID-19}}: {{Finally}},
  {{Telemedicine}} at {{Center Stage}},'' \emph{Telemedicine and e-Health},
  vol.~26, no.~9, pp. 1106--1109, May 2020.

\bibitem{sobrepera2021PerceivedUsefulnessSocial}
M.~J. Sobrepera, V.~G. Lee, S.~Garg, R.~Mendonca, and M.~J. Johnson,
  ``Perceived {{Usefulness}} of a {{Social Robot Augmented Telehealth
  Platform}} by {{Therapists}} in the {{United States}},'' \emph{IEEE Robot
  Autom Lett}, vol.~6, no.~2, pp. 2946--2953, Apr. 2021.

\bibitem{feil-seifer2005DefiningSociallyAssistive}
D.~{Feil-Seifer} and M.~J. Matari{\'c}, ``Defining {{Socially Assistive
  Robotics}},'' in \emph{9th {{Int}}. {{Conf}}. {{Rehabilitation
  Robotics}}}.\hskip 1em plus 0.5em minus 0.4em\relax {Chicago, IL, USA}:
  {IEEE}, 2005, pp. 465--468.

\bibitem{pulido2019SociallyAssistiveRobotic}
J.~C. Pulido, C.~{Suarez-Mejias} \emph{et~al.}, ``A {{Socially Assistive
  Robotic Platform}} for {{Upper-Limb Rehabilitation}}: {{A Longitudinal Study
  With Pediatric Patients}},'' \emph{IEEE Robot Autom Mag}, vol.~26, no.~2, pp.
  24--39, Jun. 2019.

\bibitem{fridin2014RoboticsAgentCoacher}
M.~Fridin and M.~Belokopytov, ``Robotics {{Agent Coacher}} for {{CP}} motor
  {{Function}} ({{RAC CP Fun}}),'' \emph{Robotica}, vol.~32, no.~8, pp.
  1265--1279, 2014.

\bibitem{fridin2014KindergartenSocialAssistive}
M.~Fridin, ``Kindergarten social assistive robot: {{First}} meeting and ethical
  issues,'' \emph{Comput Human Behav}, vol.~30, pp. 262--272, Jan. 2014.

\bibitem{fasola2013SociallyAssistiveRobot}
J.~Fasola and M.~J. Matari{\'c}, ``A {{Socially Assistive Robot Exercise
  Coach}} for the {{Elderly}},'' \emph{J Hum Robot Interact}, vol.~2, no.~2,
  pp. 3--32, 2013.

\bibitem{fridin2014EmbodiedRobotVirtual}
M.~Fridin and M.~Belokopytov, ``Embodied {{Robot}} versus {{Virtual Agent}}:
  {{Involvement}} of {{Preschool Children}} in {{Motor Task Performance}},''
  \emph{Int J Hum Comput Interact}, vol.~30, no.~6, pp. 459--469, Jun. 2014.

\bibitem{bainbridge2011BenefitsInteractionsPhysically}
W.~A. Bainbridge, J.~W. Hart, E.~S. Kim, and B.~Scassellati, ``The {{Benefits}}
  of {{Interactions}} with {{Physically Present Robots}} over {{Video-Displayed
  Agents}},'' \emph{Int J Soc Robot}, vol.~3, no.~1, pp. 41--52, Jan. 2011.

\bibitem{kiesler2008AnthropomorphicInteractionsRobot}
S.~Kiesler, A.~Powers, S.~Fussell, and C.~Torrey, ``Anthropomorphic
  {{Interactions}} with a {{Robot}} and {{Robot}}\textendash like {{Agent}},''
  \emph{Soc Cogn}, vol.~26, no.~2, pp. 169--181, 2008.

\bibitem{vasco2019TrainMeStudy}
V.~Vasco, C.~Willemse \emph{et~al.}, ``Train with {{Me}}: {{A Study Comparing}}
  a {{Socially Assistive Robot}} and a {{Virtual Agent}} for a {{Rehabilitation
  Task}},'' in \emph{Social {{Robotics}}}, ser. Lecture {{Notes}} in {{Computer
  Science}}, M.~A. Salichs, S.~S. Ge \emph{et~al.}, Eds., 2019, pp. 453--463.

\bibitem{mann2015PeopleRespondBetter}
J.~A. Mann, B.~A. Macdonald, I.-H. Kuo, X.~Li, and E.~Broadbent, ``People
  respond better to robots than computer tablets delivering healthcare
  instructions,'' \emph{Comput. Hum. Behav.}, vol.~43, pp. 112--117, 2015.

\bibitem{cespedes2020SocialHumanRobotInteraction}
N.~C{\'e}spedes, M.~M{\'u}nera, C.~G{\'o}mez, and C.~A. Cifuentes, ``Social
  {{Human-Robot Interaction}} for {{Gait Rehabilitation}},'' \emph{IEEE Trans
  Neural Syst Rehabil Eng}, vol.~28, no.~6, pp. 1299--1307, Jun. 2020.

\bibitem{adalgeirsson2010MeBotRoboticPlatform}
S.~O. Adalgeirsson and C.~Breazeal, ``{{MeBot}}: {{A Robotic Platform}} for
  {{Socially Embodied Presence}},'' in \emph{2010 5th {{ACM}}/{{IEEE Int}}.
  {{Conf}}. {{Human-Robot Interaction}}}.\hskip 1em plus 0.5em minus
  0.4em\relax {Osaka, Japan}: {IEEE}, pp. 15--22.

\bibitem{sirkin2011MotionAttentionKinetic}
D.~Sirkin, G.~Venolia \emph{et~al.}, ``Motion and {{Attention}} in a {{Kinetic
  Videoconferencing Proxy}},'' in \emph{Human-{{Computer Interaction}}
  \textendash{} {{INTERACT}} 2011}, D.~Hutchison, T.~Kanade \emph{et~al.},
  Eds.\hskip 1em plus 0.5em minus 0.4em\relax {Berlin, Heidelberg}: {Springer
  Berlin Heidelberg}, 2011, pp. 162--180.

\bibitem{dodakian2017HomeBasedTelerehabilitationProgram}
L.~Dodakian, A.~L. McKenzie \emph{et~al.}, ``A {{Home-Based Telerehabilitation
  Program}} for {{Patients}} with {{Stroke}},'' \emph{Neurorehabil Neural
  Repair}, vol.~31, no. 10-11, pp. 923--933, 2017.

\bibitem{abel2017CanTelemedicineBe}
K.~C. Abel, K.~Baldwin \emph{et~al.}, ``Can {{Telemedicine Be Used}} for
  {{Adolescent Postoperative Knee Arthroscopy Follow-up}}?'' \emph{JBJS JOPA},
  vol.~5, no.~4, pp. 1--4, Oct. 2017.

\bibitem{prvubettger2020EffectsVirtualExercise}
J.~Prvu~Bettger, C.~L. Green \emph{et~al.}, ``Effects of {{Virtual Exercise
  Rehabilitation In-Home Therapy Compared}} with {{Traditional Care After Total
  Knee Arthroplasty}}: {{VERITAS}}, a {{Randomized Controlled Trial}},''
  \emph{JBJS}, vol. 102, no.~2, pp. 101--109, Jan. 2020.

\bibitem{johnson2015CanYouSee}
S.~Johnson, I.~Rae, B.~Mutlu, and L.~Takayama, ``Can {{You See Me Now}}? {{How
  Field}} of {{View Affects Collaboration}} in {{Robotic Telepresence}},'' in
  \emph{Proc. 33rd {{Ann}}. {{ACM Conf}}. {{Human Factors}} in {{Computing
  Systems}}}.\hskip 1em plus 0.5em minus 0.4em\relax {Seoul, Korea}: {ACM},
  2015, pp. 2397--2406.

\bibitem{bettger2020COVID19MaintainingEssential}
J.~P. Bettger, A.~Thoumi \emph{et~al.}, ``{{COVID-19}}: Maintaining essential
  rehabilitation services across the care continuum,'' \emph{BMJ Glob Health},
  vol.~5, no.~5, p. e002670, May 2020.

\bibitem{sobrepera2019DesigningEvaluatingFace}
M.~J. Sobrepera, E.~Kina, and M.~J. Johnson, ``Designing and {{Evaluating}} the
  {{Face}} of {{Lil}}'{{Flo}}: {{An Affordable Social Rehabilitation Robot}},''
  in \emph{2019 {{IEEE}} 16th {{Int}}. {{Conf}}. {{Rehabilitation
  Robotics}}}.\hskip 1em plus 0.5em minus 0.4em\relax {IEEE}, pp. 748--753.

\bibitem{sobrepera2021DesignLilFlo}
M.~J. Sobrepera, V.~G. Lee, and M.~J. Johnson, ``The design of {{Lil}}'{{Flo}},
  a socially assistive robot for upper extremity motor assessment and
  rehabilitation in the community via telepresence,'' \emph{J Rehabil Assist
  Technol Eng}, vol.~8, pp. 1--26, Apr. 2021.

\bibitem{jongbloed-pereboom2013NormScoresBox}
M.~{Jongbloed-Pereboom}, M.~{Nijhuis-Van Der Sanden}, and B.~Steenbergen,
  ``Norm scores of the box and block test for children ages 3-10 years,''
  \emph{Am J Occup Ther}, vol.~67, no.~3, pp. 312--318, 2013.

\bibitem{williams1995ChildrenColorTrails}
J.~Williams, V.~Rickert \emph{et~al.}, ``Children's {{Color Trails}},''
  \emph{Arch Clin Neuropsychol}, vol.~10, no.~3, pp. 211--223, 1995.

\bibitem{maj1993EvaluationTwoNew}
M.~Maj, L.~D'Elia \emph{et~al.}, ``Evaluation of two new neuropsychological
  tests designed to minimize cultural bias in the assessment of {{HIV-1}}
  seropositive persons: A {{WHO}} study,'' \emph{Arch Clin Neuropsychol},
  vol.~8, no.~2, pp. 123--135, Mar. 1993.

\bibitem{mathiowetz1986GripPinchStrength}
V.~Mathiowetz, D.~Wiemer, and S.~Federman, ``Grip and pinch strength: Norms for
  6- to 19-year-olds.'' \emph{Am J Occup Ther}, vol.~40, no.~10, pp. 705--711,
  1986.

\bibitem{hart1988DevelopmentNASATLXTask}
S.~G. Hart and L.~E. Staveland, ``Development of {{NASA-TLX}} ({{Task Load
  Index}}): {{Results}} of {{Empirical}} and {{Theoretical Research}},'' in
  \emph{Advances in {{Psychology}}}.\hskip 1em plus 0.5em minus 0.4em\relax
  {Elsevier}, 1988, vol.~52, pp. 139--183.

\bibitem{mcauley1989PsychometricPropertiesIntrinsic}
E.~McAuley, T.~Duncan, and V.~V. Tammen, ``Psychometric properties of the
  {{Intrinsic Motivation Inventory}} in a competitive sport setting: A
  confirmatory factor analysis,'' \emph{Res Q Exerc Sport}, vol.~60, no.~1, pp.
  48--58, Mar. 1989.

\bibitem{bakken2006DevelopmentValidationUse}
S.~Bakken, L.~{Grullon-Figueroa} \emph{et~al.}, ``Development, {{Validation}},
  and {{Use}} of {{English}} and {{Spanish Versions}} of the {{Telemedicine
  Satisfaction}} and {{Usefulness Questionnaire}},'' \emph{J Am Med Inform
  Assoc}, vol.~13, no.~6, pp. 660--667, 2006.

\bibitem{mathiowetz1985AdultNormsBox}
V.~Mathiowetz, G.~Volland, N.~Kashman, and K.~Weber, ``Adult {{Norms}} for the
  {{Box}} and {{Block Test}} of {{Manual Dexterity}}.'' \emph{Am J Occup Ther},
  vol.~39, no.~6, pp. 386--391, Jun. 1985.

\end{thebibliography}

\end{document}